\documentclass[10pt,twocolumn,letterpaper]{article}

\usepackage[pagenumbers]{iccv} 
\usepackage{multirow}
\usepackage{colortbl}
\usepackage{bbding}
\usepackage{pifont} 
\usepackage{appendix}
\definecolor{iccvblue}{rgb}{0.21,0.49,0.74}
\usepackage[pagebackref,breaklinks,colorlinks,allcolors=iccvblue]{hyperref}

\def\paperID{4903} 
\def\confName{ICCV}
\def\confYear{2025}

\def\model{LightGen}

\title{\model: Efficient Image Generation through Knowledge Distillation and Direct Preference Optimization}

\author{%
  Xianfeng Wu\textsuperscript{1,2*}~~ 
  Yajing Bai\textsuperscript{1,2*}~~ 
  Haoze Zheng\textsuperscript{1,2*}~~ 
  Harold Haodong Chen\textsuperscript{1,2*}~~ 
  Yexin Liu\textsuperscript{1,2*}~~ \\
  Zihao Wang\textsuperscript{1,2}~~ 
  Xuran Ma\textsuperscript{1,2}~~ 
  Wen-Jie Shu\textsuperscript{1,2}~~ 
  Xianzu Wu\textsuperscript{1,2}~~ 
  Harry Yang\textsuperscript{1,2$\dagger$}~~ 
  Ser-Nam Lim\textsuperscript{2,3$\dagger$}~~ \\
  \\
  \textsuperscript{1}The Hong Kong University of Science and Technology, \textsuperscript{2}Everlyn AI, \textsuperscript{3}University of Central Florida \\
  \textsuperscript{*}Core Contributor, \textsuperscript{$\dagger$}Corresponding Author
}

\begin{document}


\twocolumn[{
\renewcommand\twocolumn[1][t!]{#1}%
\maketitle

\begin{center}
    \centering
    \vspace{-10pt}
    \includegraphics[width=\textwidth]{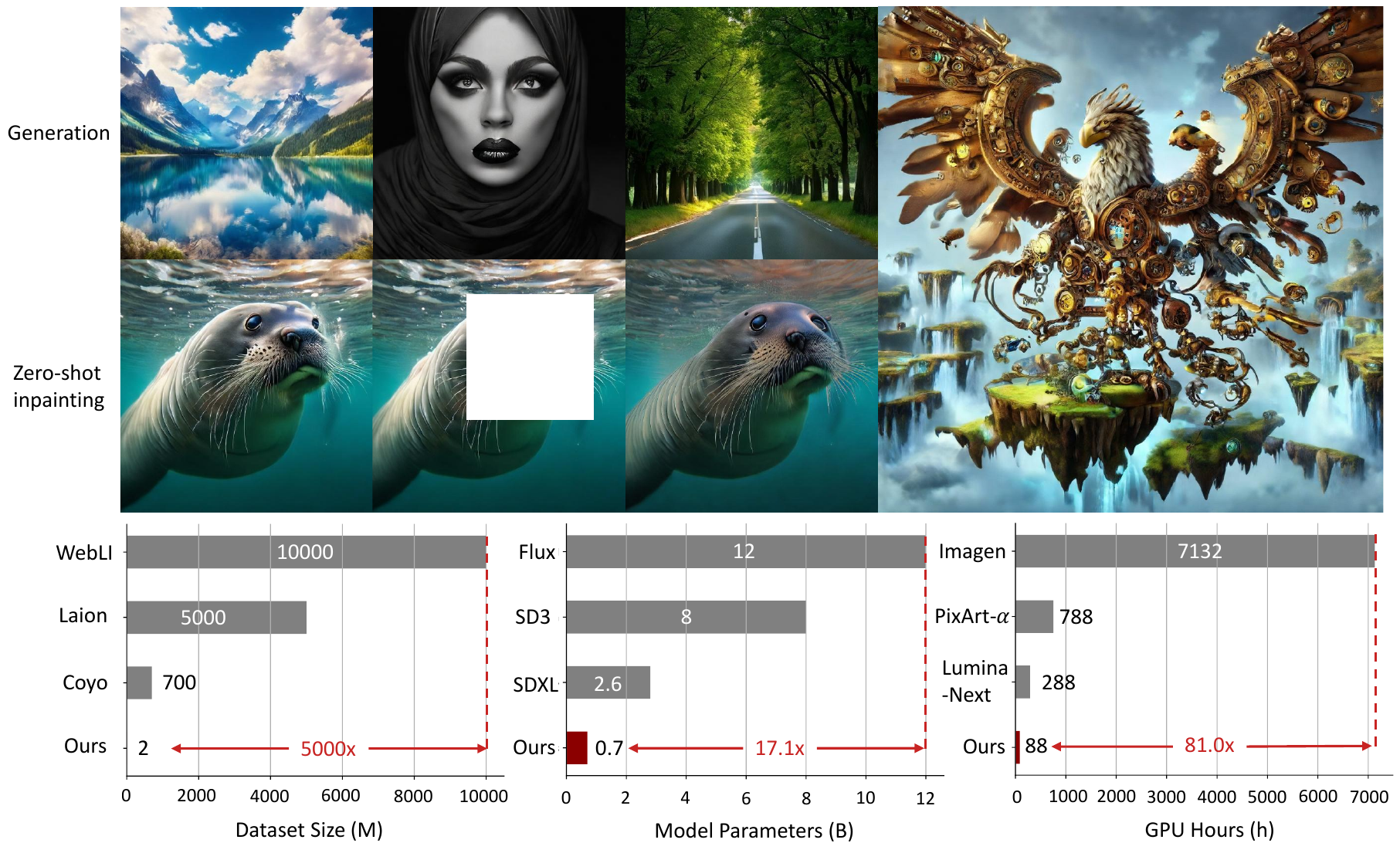}
    \vspace{-18pt}
        \captionof{figure}{Overview of \model’s capabilities in image generation, zero-shot inpainting, and resource usage. (\textbf{\textit{First Row}}) Images generated at multiple resolutions ($512\times512$ and $1024\times1024$) illustrate the scalability of \model. (\textbf{\textit{Second Row}}) Zero-shot inpainting results showcasing \model's inherent editing ability. (\textbf{\textit{Third Row}}) \model's resource consumption with drastically reduced dataset size, model parameters, and GPU hours compared to state-of-the-art models, demonstrates significant cost reductions without sacrificing performance.}
    \label{fig:fig1}
\end{center}}]

\begin{abstract}
Recent advances in text-to-image generation have primarily relied on extensive datasets and parameter-heavy architectures. These requirements severely limit accessibility for researchers and practitioners who lack substantial computational resources. In this paper, we introduce \model, an efficient training paradigm for image generation models that uses knowledge distillation (KD) and Direct Preference Optimization (DPO). Drawing inspiration from the success of data KD techniques widely adopted in Multi-Modal Large Language Models (MLLMs), LightGen distills knowledge from state-of-the-art (SOTA) text-to-image models into a compact Masked Autoregressive (MAR) architecture with only $0.7B$ parameters. Using a compact synthetic dataset of just $2M$ high-quality images generated from varied captions, we demonstrate that data diversity significantly outweighs data volume in determining model performance. This strategy dramatically reduces computational demands and reduces pre-training time from potentially thousands of GPU-days to merely 88 GPU-days. Furthermore, to address the inherent shortcomings of synthetic data, particularly poor high-frequency details and spatial inaccuracies, we integrate the DPO technique that refines image fidelity and positional accuracy. Comprehensive experiments confirm that LightGen achieves image generation quality comparable to SOTA models while significantly reducing computational resources and expanding accessibility for resource-constrained environments. Code is available at~\url{https://github.com/XianfengWu01/LightGen}

\end{abstract}    
\vspace{-0.6em}
\section{Introduction}
\label{sec:intro}

Text-to-image generation has witnessed significant advancements in recent years, with various generative models \cite{goodfellow2014gan, Rezende2015pmlr, zhai2023towards}, including diffusion models \cite{Rombach2022CVPR, Patrick2024ICML, flux2024, xiao2024omnigen} and autoregressive models \cite{Li2024mar, fan2025scaling}, achieving remarkable results \cite{dhariwal2021diffusion, Yang_2017_CVPR}. These models have demonstrated the ability to generate high-quality images under diverse input conditions, such as text prompts, sketches, and lighting specifications \cite{William2023ICCV, Patrick2024ICML, flux2024}. However, their reliance on large-scale datasets and parameter-heavy architectures imposes substantial costs in terms of data collection, training resources, and computational infrastructure. This makes these models less accessible to researchers and practitioners with limited access to high-performance GPU clusters or massive labeled datasets.

Although recent work has explored \textit{efficient pretraining} strategies for label-to-image generation models~\cite{yao2024fasterdit}, the development of lightweight and efficient training paradigms for text-to-image generation models remains largely underexplored. Text-to-image generation models introduce unique challenges due to embedding captions to visual tokens, it is hard to combine discrete tokens and continuous tokens, which inherently requires balancing model efficiency and performance. Moreover, the SOTA in text-to-image generation often demands $\mathcal{O}(100M)$ to $\mathcal{O}(1B)$ of labeled images and extensive pretraining on high-performance GPU clusters. These requirements are prohibitively expensive and limit the accessibility of AR models for broader research and application.

In this work, we propose \model, a novel strategy for efficient pre-training of the MAR-based text-to-image generation model. We find that achieving comparable text-to-image generation models does not necessarily require \textbf{massive datasets} and \textbf{large parameters}, and data diversity plays a more vital role than sheer data volume.
Drawing on the success of data distillation in the MLLM~\cite{deepseekai2025, he2024bunny}, we utilize high-quality image data as pre-train data come from SOTA models generate, LightGen rapidly gets a comparable result in the $256px$ pre-train stage by learning SOTA models how to generate high-quality images.
To this end, we propose leveraging synthetic datasets which use understanding caption data as the caption to ensure sufficient caption diversity distribution, then use SOTA text-to-image models to generate images as pre-train data to ensure image data are high-quality. By combining this strategy with a lightweight MAR architecture, LightGen significantly reduces the resource requirements of text-to-image generation models without sacrificing performance. As shown in~\cref{fig:fig1}, LightGen achieves efficiency in three key aspects:
\begin{itemize}[leftmargin=*]
    \item[$\blacktriangleright$] \textbf{Data Efficiency:} We only use $2M$ image data, compared to the size of pre-train data of SOTA models, we significantly reduce the data scale. ($\mathcal{O}(100M) \rightarrow \mathcal{O}(1M)$).
    \item[$\blacktriangleright$] \textbf{Parameter Efficiency:} Compare with SOTA models using $2.6B$, $8B$, $12B$ and even larger parameters, we only use the $0.7B$ model to achieve a comparable performance with SOTA models.
    \item[$\blacktriangleright$] \textbf{Training Efficiency:} In the last text-to-image pre-traing stage, they often required at least 288 A100 GPU Days or more, even tens of thousands GPU Days, we only require 88 A100 GPU Days.
\end{itemize}

However, while synthetic data addresses the challenge of data diversity, it also introduces two key limitations: \ding{182}~\textbf{Poor high-frequency information}, as synthetic data often lack fine-grained details present in real-world images; and \ding{183}~\textbf{Difficulty in position capture}, we utilize data augmentation in the pre-train stage, which makes LightGen have misalignment or simplified spatial relationships problem. To overcome these issues, we incorporate a post-processing technique called DPO (Direct Preference Optimization). DPO refines high-frequency details and enhances positional accuracy, effectively addressing the shortcomings of synthetic data. Its integration into the training pipeline further improves \model's robustness and image quality under data-scale-constrained scenarios. Extensive experiments demonstrate that LightGen achieves performance comparable to SOTA image generation models while significantly reducing resource demands and training time.
In summary, our contributions are as follows:
\begin{itemize}[leftmargin=*]
    \item \textbf{\textit{Efficient Text-to-Image Training Pipeline:}} We propose LightGen, a novel training framework for T2I generation models that significantly reduces the dependence on large-scale datasets and computational resources.
    \item \textbf{\textit{Synthetic Data Utilization:}} We find that data diversity, rather than sheer data volume, is critical for the performance of text-to-image models. Using diverse synthetic data generated by SOTA models, we achieve data efficiency while maintaining high-quality training signals.
    \item \textbf{\textit{Lightweight Architecture Design:}} LightGen incorporates a compact architecture, improving parameter efficiency and reducing memory usage, making it suitable for resource-constrained environments.
    \item \textbf{\textit{Post-Processing with DPO:}} To address the limitations of synthetic data (\textit{e.g.}, poor high-frequency details and position capture), we introduce Direct Preference Optimization (DPO) as a post-processing technique, enhancing both robustness and image quality.
\end{itemize}
\section{Related Work}
\label{sec:related}

\makeatletter
\renewcommand\subsubsection{\@startsection{subsubsection}{3}{\z@}%
                                     {-3.25ex\@plus -1ex \@minus .2ex}%
                                     {-1em}%
                                     {\normalfont\normalsize\bfseries}}
\makeatother

\subsection{Diffusion Models}
\label{sec:DM}

Diffusion models have become a fundamental class of generative models with the development of deep learning \cite{deng2024exploring, 10613828, huang2025sefar, chen2024gaussianvton, chen2024finecliper}, particularly for tasks like image synthesis~\cite{Rombach2022CVPR, dhariwal2021diffusion}. These models operate by progressively adding noise to an image and then reversing this process to generate new samples. Their ability to model complex distributions has made them pivotal in generative modeling.

\vspace{-1.2em}
\paragraph{UNet-based Diffusion Models.}
UNet-based diffusion models leverage the UNet architecture, which employs a symmetric encoder-decoder structure with skip connections \cite{chen2024omnicreator}. This architecture is particularly effective in denoising images during the reverse diffusion process, as demonstrated by recent work~\cite{Hoogeboom2023icml, Rombach2022CVPR}. However, a key limitation of UNet-based diffusion models is their inability to scale up efficiently. While these models perform well with existing architectures, further increasing the number of model parameters to improve generative capabilities has proven to be challenging.


\vspace{-1.2em}
\paragraph{Diffusion Transformer Models.}
To address the scalability limitations of UNet-based models, Diffusion Transformer (DiT) models have been introduced~\cite{William2023ICCV, Ma2024eccv, gao2023masked, rao2024modelgrow, ji2024diffusion}. These models integrate transformer architectures into the diffusion process, enabling better scaling of parameters and leading to improved generative quality. DiT models can generate more intricate details and handle more complex inputs, outperforming UNet-based models when computational resources are available. 

However, despite their superior scalability and generative quality, DiT models struggle to embed conditional signals—such as captions or specific image attributes—into the diffusion process. To overcome this, diffusion models often rely on external control mechanisms like ControlNet~\cite{Zhang_2023_ICCV} to guide image generation with specific conditions.

\subsection{Autoregressive Models}
\label{sec:AR}

\vspace{-0.2em}
The recent success of autoregressive (AR) models in the language domain~\cite{achiam2023gpt} has led researchers to explore the possibility of using a unified model that can handle both text and visual generation tasks, so they have recently gained renewed attention in the field. AR models generate images token by token, conditioning each new token on the previously generated tokens, effectively capturing complex dependencies in natural images. 


\vspace{-1.2em}
\paragraph{Pure AR Models.}
Pure AR models~\cite{chang2022maskgit, lee2022autoregressive, luo2025openmagvit2, sun2024llamagen, yu2024randomized}, such as Llamagen~\cite{sun2024llamagen}, generate images by modeling image tokens sequentially through constant-scale tokens. VAR~\cite{tian2024visual} improves image quality by pretraining on multi-scale tokens. While these models have succeeded in image generation tasks, their scalability remains challenging. As the models grow in size, the complexity of training increases, and improvements in image quality become less pronounced. Consequently, pure AR models struggle to compete with the latest diffusion-based methods, especially in terms of generative performance at larger scales.


\vspace{-1.2em}
\paragraph{AR $+$ Diffusion Models.}
A promising direction in generative modeling combines the advantages of both AR and diffusion models~\cite{xie2025showo}. By integrating the AR framework for token generation with a diffusion process to refine these generated tokens, these hybrid models aim to improve both the quality and efficiency of image generation. For example, MAR~\cite{Li2024mar} and Fluid~\cite{fan2025scaling} combine the precision of AR token generation with refining power of diffusion processes.

These hybrid models have shown significant improvements in generative performance. The AR component provides fine-grained control over token-wise generation, while the diffusion component ensures that the overall image quality is enhanced. This combination allows the models to scale more efficiently, generating high-quality images that outperform both pure AR and pure diffusion models.
\section{Preliminary}
\label{sec:preliminary}

In this section, we introduce the foundational concepts of diffusion models and AR models in the context of image generation. These models have played critical roles in advancing the state-of-the-art in generative modeling.

\begin{figure*}[t]
    \centering
    \includegraphics[width=1.0\linewidth]{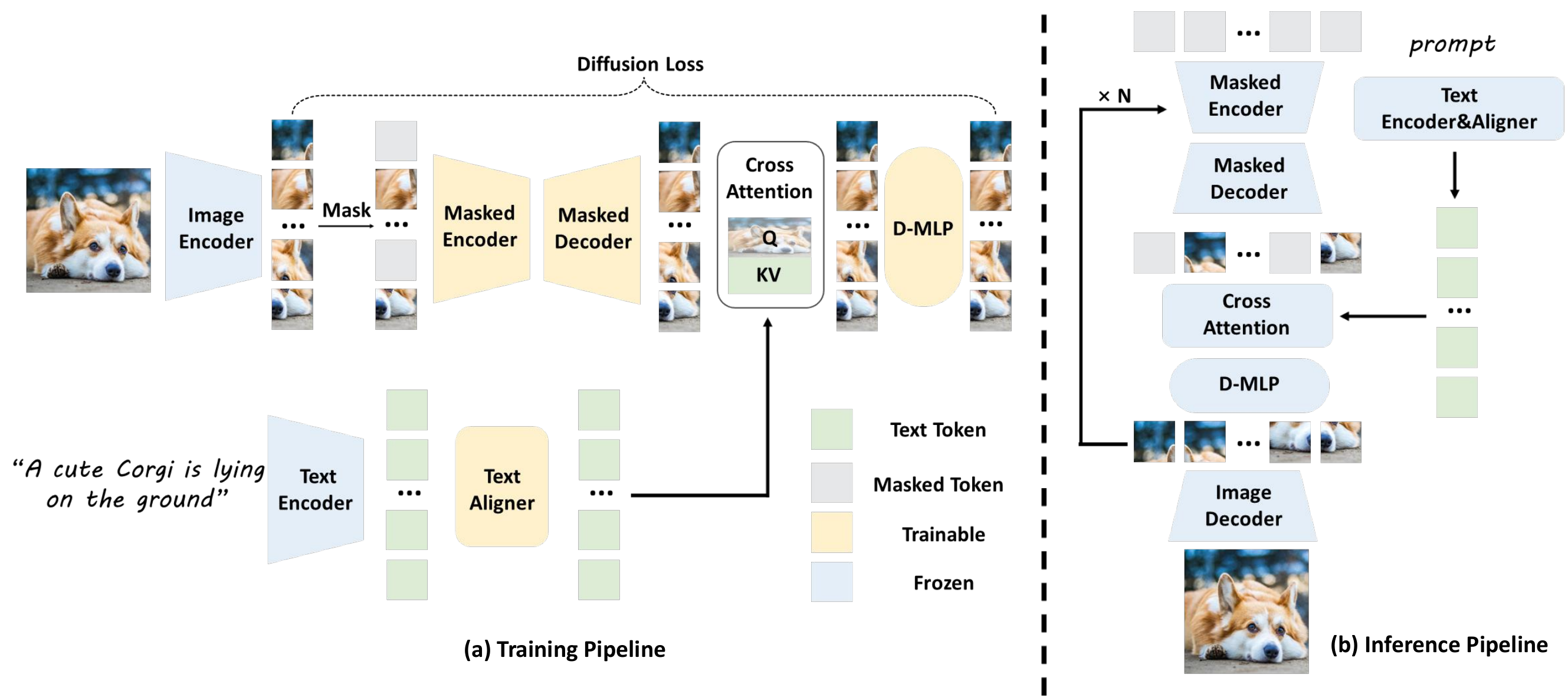}
    \vspace{-1.8em}
    \caption{Overview of LightGen efficient pretraining. (\textbf{\textit{a}}) Training: Images are encoded into tokens via a pre-trained tokenizer, while text embeddings from a T5 encoder are refined by a trainable aligner. A masked autoencoder uses text tokens as queries/values and image tokens as keys for cross-attention, followed by refinement with a Diffusion MLP (D-MLP). (\textbf{\textit{b}}) Inference: Tokens are predicted and iteratively refined over $N$ steps, then decoded by the image tokenizer to generate final images.}
    \vspace{-0.4em}
    \label{fig:fig2}
\end{figure*}

\subsection{Diffusion Models}
\label{sec:diffusion}

\paragraph{Diffusion Models.}
Diffusion models learn a process of progressively adding noise to an image and then reversing this noising process to generate new samples. The process is typically divided into two phases: a forward process, where noise is gradually added to the image, and a reverse process, where the model learns to recover the original image by denoising it. This approach has proven to be highly effective in generating high-quality images.

Given an image $\mathbf{x}_0$, the forward diffusion process defines a Markov chain that progressively adds Gaussian noise to the image, resulting in a sequence of latent variables $( \mathbf{x}_1, \mathbf{x_2}, \cdots, \mathbf{x_T} )$, where $T$ is the total number of diffusion steps. The forward process can be mathematically described as:
\setlength\abovedisplayskip{4pt}
\setlength\belowdisplayskip{4pt}
\begin{equation}
     \mathbf{x}_t = \sqrt{1 - \beta_t} \mathbf{x}_{t-1} + \beta_t \epsilon_t, \quad \epsilon_t \sim \mathcal{N}(0, I),
\label{eq:diffusion}
\end{equation}
where $\beta_t$ represents a variance schedule and $\epsilon_t$ is Gaussian noise added at each step.

The reverse diffusion process is trained to predict a clean image $\mathbf{x}_0$ from noisy observations $\mathbf{x}_t$. The model learns to approximate the reverse diffusion step:
\begin{equation}
     p(\mathbf{x}_{t-1} | \mathbf{x}_t) = \mathcal{N}(\mathbf{x}_{t-1}; \mu_\theta(\mathbf{x}_t, t), \sigma_t^2 I),
\label{eq:diffusion_reverse}
\end{equation}
where $\mu_\theta$ is the model's prediction at step $t$, and $\sigma_t^2$ is the noise variance schedule.

\vspace{-0.4em}
\subsection{Autoregressive Models}
\label{sec:autoregressive}

\vspace{-0.2em}
\subsubsection*{AR Models.}
AR models generate images by sequentially producing their components (pixels or tokens), conditioning each new element on the previously generated ones. In the context of image generation, the model produces each pixel (or token) in a sequence, where each new element depends on the ones generated before it.

Given an image $\mathbf{x} = (x_1, x_2, \cdots, x_n)$, an AR model factorizes the joint distribution of the tokens as:
\begin{equation}
     p(\mathbf{x}) = \prod_{i=1}^{n} p(x_i | x_1, x_2, \cdots, x_{i-1}),
\label{eq:ar}
\end{equation}
where $p(x_i | x_1, \cdots, x_{i-1})$ represents the conditional distribution of each token given its predecessors. This factorization allows the model to capture complex dependencies between pixels, resulting in high-quality images.

\vspace{-1.2em}
\subsubsection*{Random Autoregressive Models.}
Random autoregressive (RAR) models~\cite{yu2024random} extend the basic AR approach by generating tokens in random order. By introducing randomness into the token selection process, RAR models are able to capture global dependencies more effectively compared to standard AR models, as described in~\cref{eq:ar}. The probability distribution for RAR is given by:
\begin{equation}
    p(\mathbf{x}) = \prod_{i=1}^n p(x^{j_i} \mid x^{j_1}, x^{j_2}, \cdots, x^{j_{i-1}}),
    \label{eq:RAR}
\end{equation}
where $j_i$ is a random, non-repeating index ranging from $1$ to $n$, \textbf{Supplementary Material} prove that RAR is better than Pure AR in continous tokens' filed. 


\vspace{-1.2em}
\subsubsection*{Masked Autoregressive Models.}
Masked autoregressive models (MAR)~\cite{Li2024mar, fan2025scaling} generalize the RAR framework by predicting a set of tokens during each inference step. This formulation strikes a balance between the speed of inference and the generative capability. The probability distribution in MAR is given by:
\begin{equation}
\begin{split}
    p(\mathbf{x}) = & \prod_{i=1}^{\frac{n}{\tau}} p(x^{\tau (i-1)}, x^{\tau (i-1) + 1}, \cdots, \\
    & x^{\tau i - 1} \mid x^{j_1}, x^{j_2}, \cdots, x^{\tau (i-1) - 1}),
\label{eq:MAR}
\end{split}
\end{equation}
where $\tau$ represents the number of tokens predicted in one inference step, \textbf{Supplementary Material} has prove why MAR is better than RAR.

\section{Methodology}
\label{sec:method}

In this section, we detail the methodology used in this work, focusing on our proposed model architecture, training pipeline, and post-processing techniques. LightGen uses a novel pre-training strategy without modifying the architecture to achieve comparable generative capabilities with few datasets and GPU resources.

 
\subsection{LightGen Pipeline}
\label{sec:pipeline}

In this paper, LightGen build upon Fluid's architecture~\cite{fan2025scaling}, which serves as the base model for our approach. We also incorporate interpolated positional embeddings inspired by DINO~\cite{Caron_2021_ICCV, oquab2024dinov} to allow LightGen to generate images at different resolutions. This flexibility is crucial for training on multiple image scales and achieving high-quality image generation.

\cref{fig:fig2} (a) illustrates our training pipeline. The ground truth image is first passed through an image VAE encoder to obtain a latent representation of the image. The latent features are then patchified into continuous tokens $\mathbf{I} = \{i_1, i_2, \cdots, i_n\}$, which are subsequently masked in a random order at a high ratio. These masked tokens are input into a masked encoder-decoder architecture. The masked decoder generates token predictions based on the masked input, which are used as queries in a cross-attention mechanism.

For conditioning the model, we introduce text as an additional modality. We process the text using a T5 encoder~\cite{ni2022sentence} and align it with the image features through a text aligner. The processed text is then converted into discrete tokens $\mathbf{T} = \{t_1, t_2, \cdots, t_n\}$, which serve as the keys and values in the cross-attention mechanism. This enables the model to generate semantic tokens $\mathbf{S} = \{s_1, s_2, \cdots, s_n\}$.

These semantic tokens $\mathbf{S}$ are passed through a tiny diffusion MLP, denoted as $D(\cdot)$, which conditions the token generation process and generates high-quality image tokens $\widehat{\mathbf{I}} = \{\widehat{i}_1, \widehat{i}_2, \cdots, \widehat{i}_n\}$. These generated image tokens are then conditioned on the ground truth latent representation $\mathbf{I}$ produced by the image VAE encoder. The model parameters are updated using the diffusion loss function:
\begin{equation}
     \mathcal{L}_\theta(\mathbf{I}, \mathbf{S}) = \frac{\sum_{i=1}^n \mathbb{E}_{\epsilon, t} \left[ \left\Vert \epsilon - \epsilon_\theta(i_{i_t}|t, s_i) \right\Vert^2 \right]}{n},
\label{eq:diffusion_loss}
\end{equation}
where $\epsilon \in \mathbb{R}^d$ is a noise vector sampled from $\mathcal{N}(0, I)$. The noise-corrupted vector $i_t$ is computed via~\cref{eq:diffusion}. The variable $t$ denotes the time step of the noise schedule. The noise estimator $\epsilon_\theta$, parameterized by $\theta$, is a small MLP network, which takes $i_t$ as input and is conditional on both $t$ and $s_i$. The notation $\epsilon_\theta(i_{i_t}|t, s_i)$ indicates that the network takes $i_t$ as input, conditioned on the time step $t$ and the corresponding semantic token $s_i$.

\cref{fig:fig2} (b) shows the inference pipeline. The process begins by inputting noise tokens $\mathbf{Z} = \{z_1, z_2, \cdots, z_n\}$ of the same size into the masked encoder-decoder. The cross-attention blocks then combine the text’s discrete tokens $\mathbf{T}$ with the visual tokens. The tiny diffusion MLP $D(\cdot)$ iteratively refines the tokens, acting as a denoising function to guide the tokens towards a high-quality image representation. This process is repeated for $N$ steps to further refine the tokens. Finally, the tokens $\widehat{\mathbf{I}}$ are reshaped and passed through the image VAE decoder to produce the final image.

\subsection{Post-processing}
\label{sec:postprocessing}

\begin{table*}[t]
\centering
\vspace{-0.4em}
\caption{Performance comparison in $256\times256$ on GenEval~\cite{ghosh2023geneval}. Best results are shown in \textbf{bold}.}
\vspace{-0.8em}
\label{tab:system_comparison_256}
\resizebox{\linewidth}{!}{
\begin{tabular}{lccccccccc}
\hlineB{2.5}
\multirow{2}{*}{\textbf{Model}} & 
\multirow{2}{*}{\textbf{\#Params}} &
\multirow{2}{*}{\textbf{Pre-train Data}} &
\multicolumn{7}{c}{\textbf{GenEval}} \\
\cmidrule(lr){4-10}
 & &  & Single Obj. & Two Obj. & Colors & Counting & Position & Color Attri.  & \textbf{Overall} \\
\hlineB{2}
\rowcolor{lightgray!20} \multicolumn{10}{l}{{\textit{Diffusion-based}}} \\
Stable Diffusion v1.5~\cite{Rombach2022CVPR} & 0.9B & 2B & 0.41 & 0.05 & 0.30 & 0.02 & 0.01 & 0.01 & 0.13 \\
Stable Diffusion v2.1~\cite{Rombach2022CVPR} & 0.9B & 5B & 0.11 & 0.01 & 0.04 & 0.01 & 0.00 & 0.00 & 0.02 \\
Stable Diffusion XL~\cite{Rombach2022CVPR} & 2.6B & - & 0.05 & 0.03 & 0.30 & 0.01 & 0.00 & 0.00 & 0.01 \\
Stable Diffusion 3~\cite{Patrick2024ICML} & 8B & - & 0.89 & \textbf{0.62} & 0.70 & 0.30 & \textbf{0.17} & 0.33 & 0.50 \\
Flux~\cite{flux2024} & - & 12B & 0.96 & 0.58 & 0.77 & \textbf{0.48} & 0.13 & \textbf{0.33} & \textbf{0.54} \\
\hline
\rowcolor{lightgray!20} \multicolumn{10}{l}{{\textit{Autoregressive}}} \\
Llamagen~\cite{sun2024llamagen} & 0.7B & 50M & 0.69 & 0.34 & 0.55 & 0.19 & 0.06 & 0.02 & 0.31 \\
\hdashline
\rowcolor{yellow!20}
\textbf{LightGen w/o DPO (80k steps)} & 0.7B & 2M & 0.98 & 0.44 & \textbf{0.85} & 0.36 & 0.08 & 0.24 & 0.49 \\
\rowcolor{yellow!20}
\textbf{LightGen w/o DPO} & 0.7B & 2M & \textbf{0.99} & 0.53 & 0.85 & 0.40 & 0.13 & 0.25 & 0.53 \\
\hlineB{2.5}
\end{tabular}
}
\end{table*}
\begin{table*}[t]
\centering
\vspace{-0.4em}
\caption{Performance comparison in $512\times512$ on GenEval~\cite{ghosh2023geneval}.}
\vspace{-0.8em}
\label{tab:system_comparison_512}
\resizebox{\linewidth}{!}{
\begin{tabular}{lccccccccc}
\hlineB{2.5}
\multirow{2}{*}{\textbf{Model}} & 
\multirow{2}{*}{\textbf{\#Params}} &
\multirow{2}{*}{\textbf{Pre-train Data}} &
\multicolumn{7}{c}{\textbf{GenEval}} \\
\cmidrule(lr){4-10}
 & &  & Single Obj. & Two Obj. & Colors & Counting & Position & Color Attri.  & \textbf{Overall} \\
\hlineB{2}
\rowcolor{lightgray!20} \multicolumn{10}{l}{{\textit{Diffusion-based}}} \\
Stable Diffusion v1.5~\cite{Rombach2022CVPR} & 0.9B & 2B & 0.96 & 0.38 & 0.77 & 0.37 & 0.03 & 0.05 & 0.42 \\
Stable Diffusion v2.1~\cite{Rombach2022CVPR} & 0.9B & 2B & 0.91 & 0.24 & 0.69 & 0.14 & 0.03 & 0.06 & 0.34 \\
Stable Diffusion XL~\cite{Rombach2022CVPR} & 2.6B & 5B & 0.63 & 0.23 & 0.51 & 0.12 & 0.04  & 0.05 & 0.26 \\
Stable Diffusion 3~\cite{Patrick2024ICML} & 8B & - & 0.99 & \textbf{0.82} & 0.80 & 0.51 & \textbf{0.27} & \textbf{0.52} & \textbf{0.65} \\
Flux~\cite{flux2024}  & 12B & - & 0.97 & 0.71 & \textbf{0.68} & 0.76 & 0.15 & 0.43 & 0.61 \\
\hline
\rowcolor{lightgray!20} \multicolumn{10}{l}{{\textit{Autoregressive}}} \\
Llamagen~\cite{sun2024llamagen} & 0.7B & 50M & 0.19 & 0.16 & 0.10 & 0.03 & 0.09 & 0.01 & 0.10 \\
\hdashline
\rowcolor{yellow!20}
\textbf{LightGen w/o DPO} & 0.7B & 2M & 0.98 & 0.58 & 0.86 & 0.37 & 0.14 & 0.28 & 0.53 \\
\rowcolor{cyan!10}
\textbf{\model} & 0.7B & 2M & \textbf{0.99} & 0.65 & \textbf{0.87} & 0.60 & 0.22 & 0.43 & 0.62 \\
\hlineB{2.5}
\end{tabular}
}
\vspace{-0.8em}
\end{table*}

After pre-training on images of size $256 \times 256$ and fine-tuning on $512 \times 512$ images, we observe that the model achieves competitive image generation performance. However, we identify several potential issues with the generator. Specifically, the model was pre-trained on a dataset containing artifacts, and the preprocessing method imitate DIT~\cite{William2023ICCV} and ADM~\cite{dhariwal2021diffusion} will flip image cannot let models learning positional signals. As a result, the model may struggle to capture fine-grained spatial relationships in the generated images.

To address these concerns and improve the performance of our generator, we introduce Direct Preference Optimization (DPO)~\cite{rafailov2023direct}. DPO is a post-processing technique that fine-tunes the model by optimizing its ability to generate high-quality images while incorporating positional signals. This post-processing stage helps the model better handle spatial relationships and refine the generated images, improving both quality and coherence.

\begin{figure}
    \centering
    \includegraphics[width=1.0\linewidth]{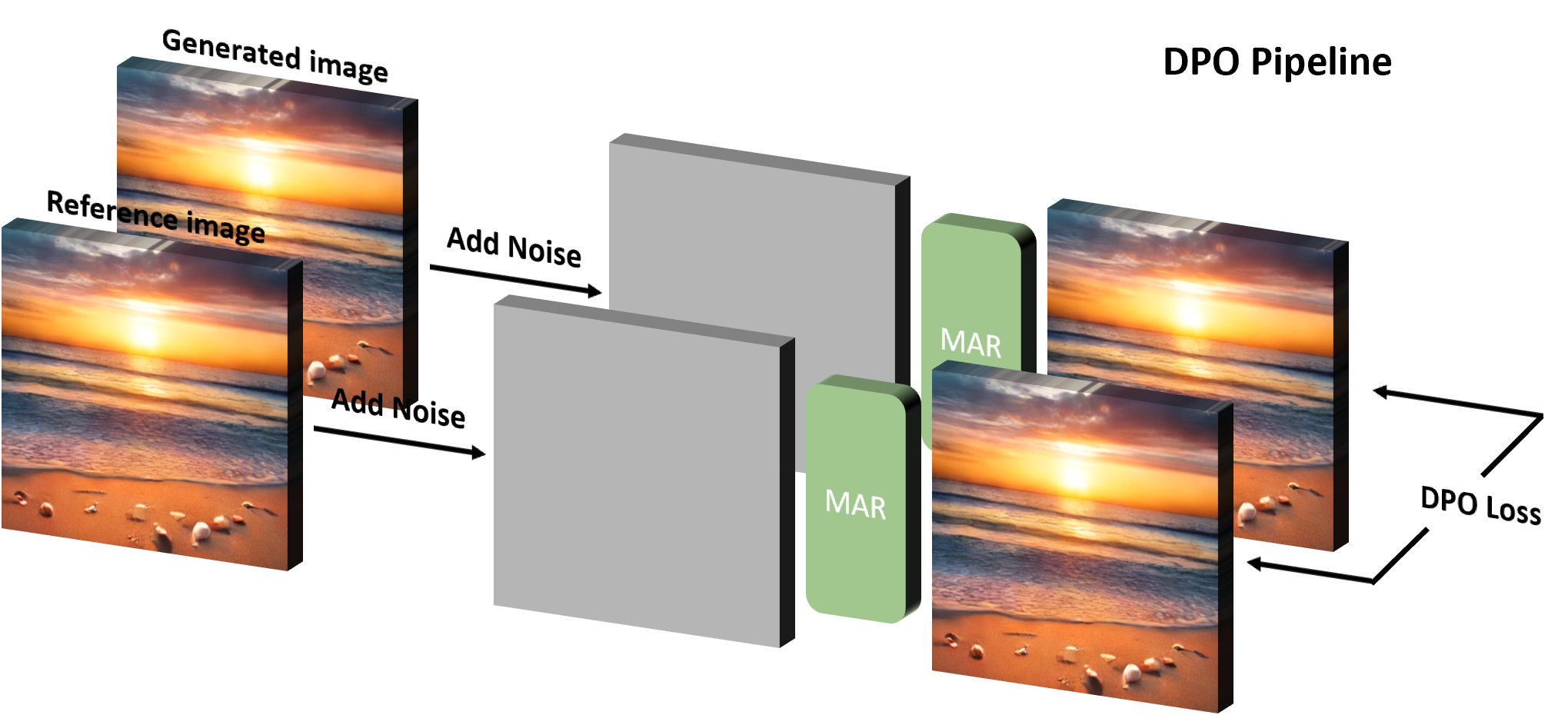}
    \caption{Illustrate of DPO Post-processing of LightGen.}	
    \vspace{-0.8em}
    \label{fig:fig3}
\end{figure}

DPO optimizes the model’s outputs by minimizing the difference between the generated and reference images under the learned preference model. Given a generated image $x_{0}^w$ and a reference image $x_{0}^l$, DPO~\cite{Wallace_2024_CVPR} aims to minimize the following loss function:
\begin{equation}
    \begin{split}
        \mathcal{L}_{DPO}(\theta) = & -\mathbb{E}_{(x_{0}^w, x_{0}^l) \sim D, t \sim \mathcal{U}(0, T)} log\sigma \Bigg\{ \\
        & -\beta T w(\lambda_t) \Bigg[\mathcal{L}_\theta(x_{t}^w, t) -  \mathcal{L}_{ref}(x_{t}^w, t)  \\
        & - (\mathcal{L}_\theta(x_{t}^l, t) - \mathcal{L}_{ref}(x_{t}^l, t)) \Bigg] \Bigg\},
    \end{split}
\label{eq:dpo_loss}
\end{equation}
where $\sigma (\cdot)$ is  sigmoid function. $x_{t}^w, x_{t}^l$ can calculate via~\cref{eq:diffusion}. $ \lambda_t = \frac{\alpha_t^2}{\sigma_t^2}$ is the signal-to-noise ratio, $w(\lambda_t)$ a weighting function. $\mathcal{L}_{\delta}(x_{t}^{\gamma}, t) =\Vert \epsilon^\gamma - \epsilon_\delta (x_{t}^\gamma, t) \Vert^2_2, \delta \in \{\theta, ref\}, \gamma \in \{w, l\},   t \sim \mathcal{U}(0, T)$.  This loss encourages $\epsilon_\theta$ to improve more at denoising $x^w_t$ than $x^l_t$. The DPO optimization process iteratively refines the generated image by updating the model parameters to align the generated output more closely with the reference image in~\cref{fig:fig3}. This results in improved generalization, allowing the model to generate higher-quality images across a wider range of tasks.

By applying DPO after the pre-training and fine-tuning stages, we refine the model to generate better quality image, and it can handle positional signals and spatial relationships.

\subsection{Data Distillation}
\label{sec:dataset_theoretical}

In contrast to SOTA image generative models use large image data and employ complex data pipelines to get large data to pretrain, LightGen leverages an artifact dataset~\cite{zk2024text2img} use understanding dataset~\cite{Dalle3_1M2024, liu2024llavanext, liu2023visual} to ensure that caption set $\mathcal{T}=\{ t_1, t_2, \cdots, t_n\}$ has enough diversity. Subsequently, SOTA image generators (e.g., DALLE3~\cite{openai2023dalle3}, Flux~\cite{flux2024}) transform semantic richness caption into a visually diverse dataset $\mathcal{D}=\{ d_1, d_2, \cdots, d_n\}$.

During both the pre-training and high-resolution stages, our target is to learn the feature representations (logits) of the artifact dataset like offline knowledge distillation~\cite{hinton2015distilling}. Instead of relying on hard targets, our model learns from the soft targets provided by SOTA models. Formally, we define our objective as:
\begin{equation}
    \mathcal{L}_\theta(\mathbf{D}) = \frac{1}{n} \sum_{i=1}^n \mathbb{E}_{\epsilon, t} \left[ \left\Vert \epsilon - \epsilon_\theta\bigl( f_{i_t} \mid t \bigr) \right\Vert^2 \right],
    \label{eq:generate}
\end{equation}
where $f_i = \mathrm{Enc}(d_i)$ denotes the feature representation (logits) extracted from the image $d_i$ (generate by SOTA models) by a image VAE encoder. $\epsilon_\theta$ is our student model parameterized by $\theta$. 

Under assumptions, the gradient of the loss with respect to the parameters $\theta$ is given by:
\begin{equation}
    \nabla_\theta \mathcal{L}_\theta(\mathbf{D}) = \frac{2}{n} \sum_{i=1}^n \mathbb{E}_{\epsilon, t} \left[ \left( \epsilon_\theta(f_{i_t} \mid t) - \epsilon \right) \nabla_\theta \epsilon_\theta(f_{i_t} \mid t) \right].
    \label{eq:total_loss}
\end{equation}
If $\epsilon_\theta$ is sufficiently expressive, then with enough diverse samples the gradient becomes an \textbf{unbiased estimator} (it will prove in \textbf{Supplementary Material}) of the true gradient of the expected loss. Hence, this method ensuring that the student model converges towards approximating the teacher’s behavior.


\begin{figure*}[t]
    \centering
    \includegraphics[width=0.95\linewidth]{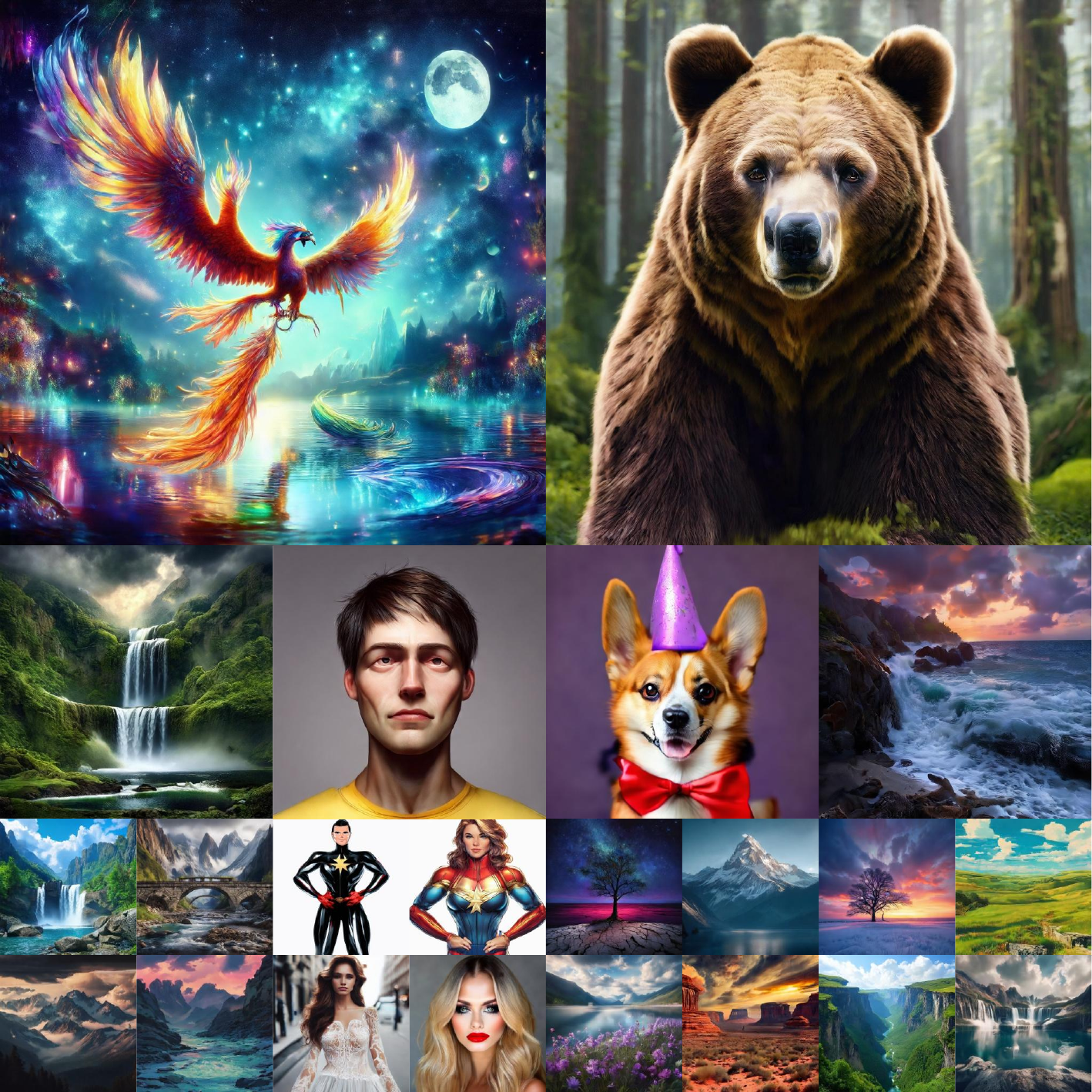}
    \vspace{-0.4em}
    \caption{\textbf{Visualization Results.} Sample outputs generated using LightGen, showcasing high-quality images at multiple resolutions ($256 \times 256$, $512 \times 512$, $1024 \times 1024$) and across diverse styles (realistic, animated, virtual, etc.), which demonstrate the versatility and scalability of our approach.}
    \label{fig:fig4}
\end{figure*}

\begin{figure*}[t]
    \centering
    \includegraphics[width=0.95\linewidth]{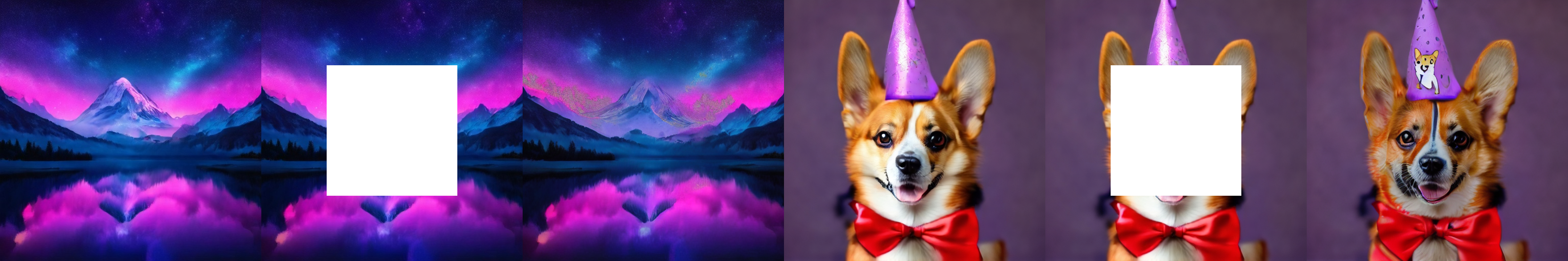}
    \vspace{-0.4em}
    \caption{Image inpainting demonstrations.}	
    \label{fig:fig5}
\end{figure*}
\section{Experiment}
\label{sec:experimental}



In this section, we present our experimental setup and results. We first describe the datasets, precomputation of image and text features, and the training/inference procedures. We then compare our method to previous systems on standard benchmarks, followed by an ablation study examining the key design choices in our approach.

\subsection{Experimental Setup}

\subsubsection*{Datasets.}
We conduct experiments on a samll high-quality artifact dataset~\cite{zk2024text2img}, which use SOTA MLLM like GPT-4o~\cite{achiam2023gpt} and Qwen2VL~\cite{wang2024qwen2vl} to generate high-quality caption, then use SOTA image generation models like Flux~\cite{flux2024} and DALLE3~\cite{openai2023dalle3} to generate high-quality image datas. To maintain consistent input dimensions, we center-crop and resize images to $256 \times 256$ for the initial pre-training stage, then fine-tune at $512 \times 512$. In addition, we evaluate on GenEval~\cite{ghosh2023geneval} and FID to measure text alignment and generation quality. For the DPO training stage, we use LAION-aesthethic~\cite{laion2022} as the DPO dataset and use Qwen2.5VL to filter out some low quality data.

\vspace{-1.2em}
\subsubsection*{Precomputation of Image and Text Features.}
To reduce redundant encoding costs and stabilize training:
\begin{itemize}
    \item \textbf{Image Features:} We use a high-quality VAE~\cite{Patrick2024ICML} to obtain latent representations of the images. Each image is encoded into a set of continuous latent tokens.
    \item \textbf{Text Features:} We employ a T5-XXL encoder~\cite{Chung2024jmlr} to convert text prompts into high-dimensional embeddings. These embeddings are then tokenized or projected to match the dimensions expected by our model.
\end{itemize}

\vspace{-1.2em}
\subsubsection*{Training \& Post-processing.}
Unless otherwise specified, we train our model using the AdamW optimizer~\cite{loshchilov2018decoupled} with $\beta_1 = 0.9$, $\beta_2 = 0.99$, and a weight decay of $0.02$. We use a linear warm-up schedule for the first few epochs, then use constant learning rate in $1e-4$. The initial pre-training is conducted for $100k$ steps at $256 \times 256$ resolution using a total batch size of 2048. For fine-tuning at $512 \times 512$, we train for an additional few steps, using a reduced learning rate of $1e-5$ ($0.1$x the base learning rate). We using same training setting in DPO training stage, and we use very samll constant learning rate $1e-8$ and $\beta$ in ~\cref{eq:dpo_loss} is $5000$.

\vspace{-1.2em}
\subsubsection*{Inference.}
During inference, we follow the pipeline described in Section~\ref{sec:method}. We use the precomputed T5-XXL text embeddings as conditioning and sample noise vectors for the image latents. We perform 64 iterative refinement steps using our masked autoregressive $+$ diffusion approach. The final latent representations are then decoded back to the pixel space via the VAE decoder.

\vspace{-1.2em}
\subsubsection*{Evaluation.}
We report the performance on standard benchmarks GenEval~\cite{ghosh2023geneval} for image quality evaluation. 

\subsection{Main Results}
\label{sec:benchmarking}

\begin{figure}[t]
    \centering
    \includegraphics[width=1.0\linewidth]{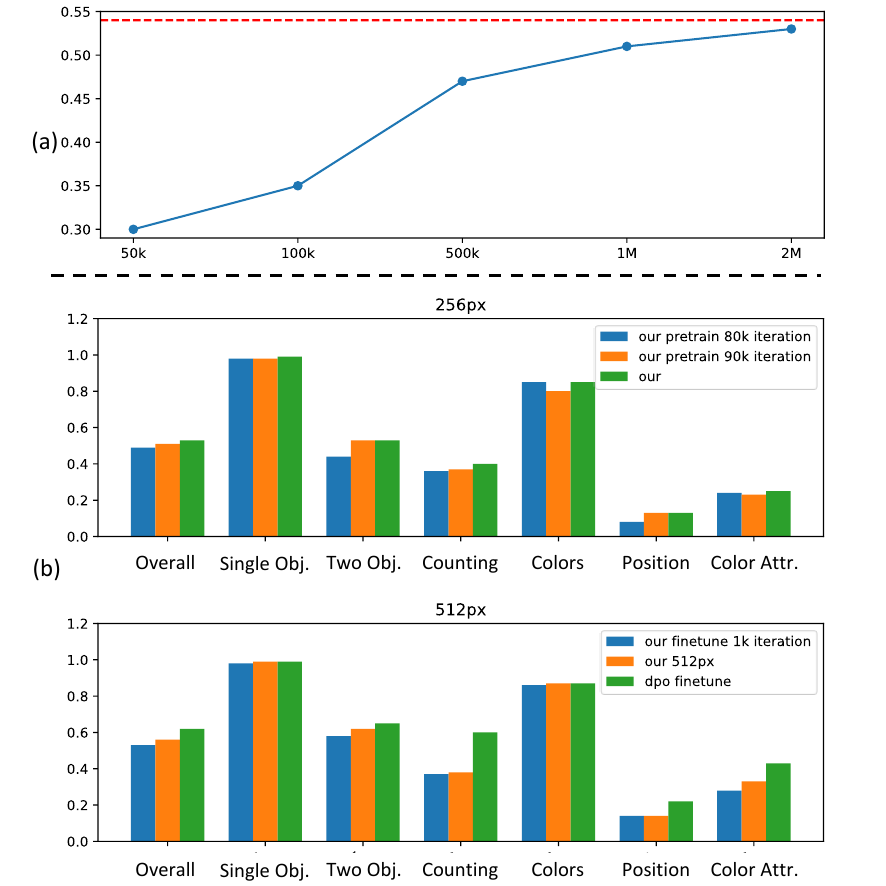}
    \vspace{-1.6em}
    \caption{Ablation studies. (\textit{\textbf{a}}) Pre-train in different data scales, we find it achieves the limitation when pre-train in $\mathcal{O}(1M)$ data scale. (\textit{\textbf{b}}) demonstrate different iteration results.}	
    \vspace{-0.8em}
    \label{fig:fig6}
\end{figure}

We compare our method against several SOTA text-to-image models, including Diffusion-based model(\textit{e.g.} Stable Diffusion~\cite{Rombach2022CVPR, Patrick2024ICML} and Flux~\cite{flux2024}), Autoregressive models (\textit{e.g.} Llamagen~\cite{sun2024llamagen}).

\vspace{-1.2em}
\subsubsection*{Quantitative Results.}
\cref{tab:system_comparison_256} and~\cref{tab:system_comparison_512} illustrate the performance of LightGen on the GenEval benchmark in resolutions of $256 \times 256$ and $512 \times 512$. At $256 \times 256$ resolution, LightGen significantly outperforms diffusion-based and autoregressive models in the task Single Object, Two Objects, and Colors, achieving overall performance scores of 0.49 (80k steps without DPO) and 0.53 (without DPO). At higher $512 \times 512$ resolution, LightGen achieves a comparable result with an overall score of 0.62, almost surpassing all SOTA models. In particular, the integration of DPO consistently enhances performance in positional accuracy and high-frequency details, underscoring its effectiveness in addressing the limitations of synthetic data.

\vspace{-1.2em}
\subsubsection*{Qualitative Results.}
\cref{fig:fig4} and~\cref{fig:fig5} visually demonstrate the capabilities and versatility of LightGen. Figure 4 shows sample outputs at multiple resolutions ($256 \times 256$, $512 \times 512$, $1024 \times 1024$) and in various artistic styles, including realistic, animated, and virtual images. \cref{fig:fig5} further illustrates the strength of LightGen through image inpainting demonstrations.

\subsection{Ablation Study}
\label{sec:ablation}


\cref{fig:fig6} (a) analyzes the impact of varying the pre-training data scale. The results indicate that performance reaches a bottleneck when the dataset scale approaches approximately 1M images, suggesting that increasing beyond this scale yields diminishing returns. Consequently, we select 2M images as our optimal pre-training dataset size. \cref{fig:fig6} (b) explores the effects of different training iterations on GenEval performance at resolutions of 256px and 512px. Notably, we observe that at the 256px stage, reasonable performance can be achieved after just 80k training iterations, highlighting the efficiency of data distillation.

\section{Conclusion}
\label{sec:conclusion}




In this paper, we propose LightGen, a novel and efficient pipeline to accelerate text-to-image generation training through KD and DPO. Our results demonstrate that efficient image generation can be realized without sacrificing quality by focusing on data diversity, compact architectures, and pre-train strategic. The proposed methodology opens avenues for exploring similar efficiency-focused training paradigms in related generative tasks such as video generation. We hope that our work encourages further research in developing accessible and efficient generative models.

{
    \small
    \bibliographystyle{ieeenat_fullname}
    \bibliography{main}
}

\appendix

\twocolumn[{
\renewcommand\twocolumn[1][t!]{#1}%
\maketitlesupplementary
\appendix

\begin{center}
    \centering
    \includegraphics[width=\textwidth]{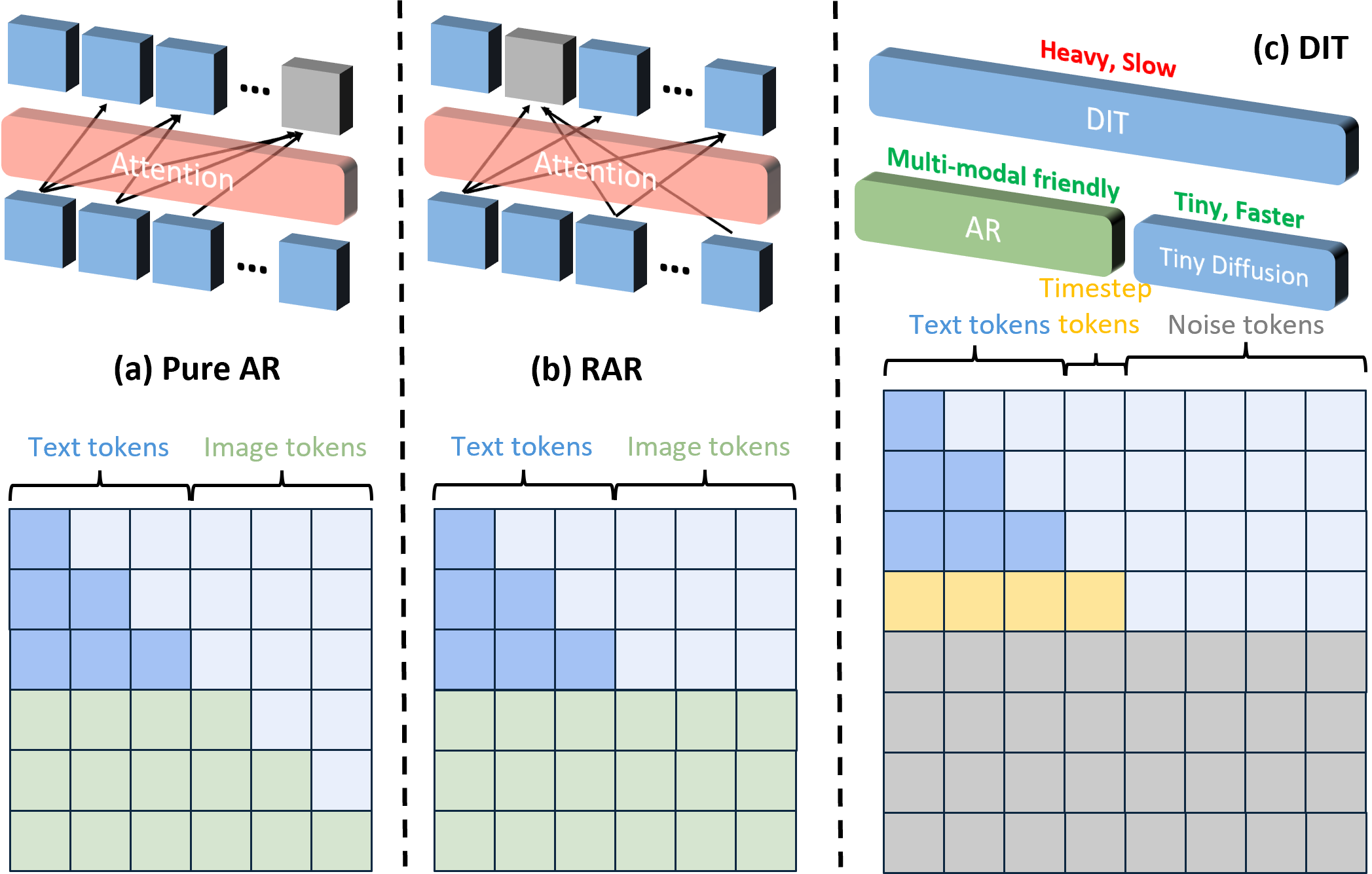}
    \vspace{-20pt}
        \captionof{figure}{Illustrate three different paradigms of image generative models.}
    \label{fig:RAR_AR}
\end{center}}]

\begin{table*}[t]
\centering
\vspace{-0.4em}
\caption{Performance comparison with other work's paper demonstrate on GenEval~\cite{ghosh2023geneval}.}
\vspace{-0.8em}
\label{tab:system_comparison}
\resizebox{\linewidth}{!}{
\begin{tabular}{lccccccccc}
\hlineB{2.5}
\multirow{2}{*}{\textbf{Model}} & 
\multirow{2}{*}{\textbf{\#Params}} &
\multirow{2}{*}{\textbf{Pre-train Data}} &
\multicolumn{7}{c}{\textbf{GenEval}} \\
\cmidrule(lr){4-10}
 & &  & Single Obj. & Two Obj. & Colors & Counting & Position & Color Attri.  & \textbf{Overall} \\
\hlineB{2}
\rowcolor{lightgray!20} \multicolumn{10}{l}{{\textit{Diffusion-based}}} \\
Stable Diffusion v1.5~\cite{Rombach2022CVPR} & 0.9B & 2B & 0.97 & 0.38 & 0.76 & 0.35 & 0.04 & 0.06 & 0.43 \\
Stable Diffusion v2.1~\cite{Rombach2022CVPR} & 0.9B & 2B & 0.98 & 0.51 & 0.85 & 0.44 & 0.07 & 0.17 & 0.50 \\
Stable Diffusion XL~\cite{Rombach2022CVPR} & 2.6B & 5B & 0.98 & 0.74 & 0.85 & 0.39 & 0.15 & 0.23 & 0.55 \\
DALLE 3~\cite{openai2023dalle3} & - & - & 0.96 & 0.87 & 0.83 & 0.47 & \textbf{0.43} & 0.45 & 0.67 \\
Stable Diffusion 3~\cite{Patrick2024ICML} & 8B & - & 0.98 & 0.84 & 0.74 & \textbf{0.66} & 0.40 & 0.43 & \textbf{0.68} \\
Flux~\cite{flux2024}  & 12B & - & 0.98 & 0.81 & 0.79 & 0.74 & 0.22 & 0.45 & 0.66 \\
\hline
\rowcolor{lightgray!20} \multicolumn{10}{l}{{\textit{Autoregressive}}} \\
Llamagen~\cite{sun2024llamagen} & 0.7B & 50M & 0.19 & 0.16 & 0.10 & 0.03 & 0.09 & 0.01 & 0.10 \\
Chameleon~\cite{chameleonteam2024} & 7B & - & - & - & - & - & - & - & 0.39 \\
SEED-X~\cite{ge2025seedx} & 17B & - & 0.96 & 0.65 & 0.80 & 0.31 & 0.18 & 0.14 & 0.51 \\
Show-o~\cite{xie2025showo} & 1.3B & - & 0.95 & 0.52 & 0.82 & 0.49 & 0.11 & 0.28 & 0.53 \\
\hdashline
\rowcolor{yellow!20}
\rowcolor{cyan!10}
\textbf{\model} & 0.7B & 2M & \textbf{0.99} & 0.65 & \textbf{0.87} & 0.60 & 0.22 & 0.43 & 0.62 \\
\hlineB{2.5}
\end{tabular}
}
\vspace{-0.8em}
\end{table*}

\begin{figure*}[t]
    \centering
    \includegraphics[width=0.95\linewidth]{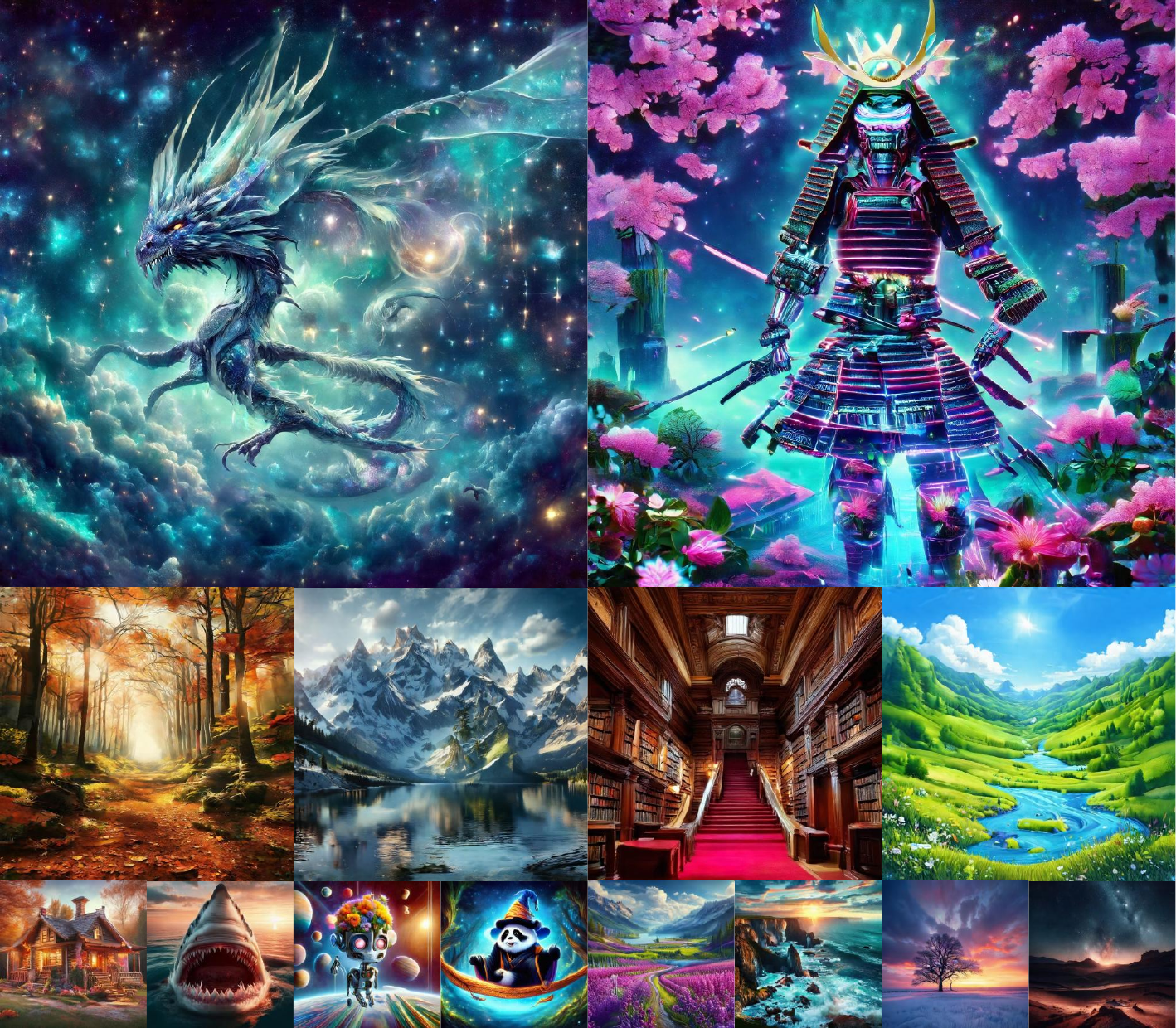}
    \caption{Additional generation result.}	
    \label{fig:fig_appen}
\end{figure*}

\section{RAR is better than Pure AR}

In the traditional deep learning field~\cite{wu2024fsc, wu2022point, luan2023high}, especially MLLM, Pure AR achieves great success. The probability distribution for RAR models~\cite{yu2024randomized}, as illustrated in~\cref{fig:RAR_AR} (b), is defined as:
\begin{equation}
p(\mathbf{x}) = \prod_{i=1}^n p(x^{j_i} \mid x^{j_1}, x^{j_2}, \cdots, x^{j_{i-1}}).
\label{eq:RAR}
\end{equation}

During training, the distribution in RAR can be approximated by:
\begin{equation}
p_{R_{\theta}}(\cdot) \approx \prod_{i=1}^{n} p(x^i \mid x^{S \backslash i}),
\label{eq:RAR_1}
\end{equation}
where $S = [1, n]$ represents the set of all token indices, and $x^{S \backslash i}$ denotes all tokens except the current token $x^i$. In this formulation, each token $x^i$ is inferred using the global context provided by all other tokens. This global conditioning enables RAR models to capture richer inter-token dependencies compared to traditional AR models, which rely solely on a sequential left-to-right generation process.

The continuous representations utilized in DIT have demonstrated exceptional generative capabilities in the visual generation domain, suggesting the inherent preference for continuous and global context in visual content generation, such as images and videos. RAR employs a bidirectional inference approach analogous to continuous representation in diffusion models, thereby leveraging global contextual information. This bidirectional conditioning enables RAR models to better approximate the superior generative performance of diffusion-based methods compared to pure AR models, leading to improved image fidelity and consistency.

\section{Why MAR is better?}

The probability distribution in MAR~\cite{Li2024mar, fan2025scaling} in~\cref{fig:RAR_AR} (c) is given by:
\begin{equation}
\begin{split}
    p(\mathbf{x}) = & \prod_{i=1}^{\frac{n}{\tau}} p(x^{\tau (i-1)}, x^{\tau (i-1) + 1}, \cdots, \\
    & x^{\tau i - 1} \mid x^{j_1}, x^{j_2}, \cdots, x^{\tau (i-1) - 1}),
\label{eq:MAR}
\end{split}
\end{equation}
where $\tau$ represents the number of tokens predicted in one inference step

When $\tau = 1$, MAR reduces to the RAR model. By increasing $\tau$, MAR offers a trade-off between:
\begin{itemize}
    \item \textbf{Performance:} Smaller values of $\tau$ capture fine-grained dependencies between tokens, improving the quality of image generation.
    \item \textbf{Efficiency:} Larger values of $\tau$ speed up inference by enabling parallelization of token predictions, reducing computation time.
\end{itemize}

This balance allows MAR models to achieve better performance and efficiency compared to both standard AR and RAR models.

\section{Unbiased Estimator}

To see why the gradient
\begin{equation}
    \nabla_\theta \mathcal{L}_\theta(\mathbf{D}) = \frac{2}{n} \sum_{i=1}^n \mathbb{E}_{\epsilon, t} \left[ \left( \epsilon_\theta(f_{i_t} \mid t) - \epsilon \right) \nabla_\theta \epsilon_\theta(f_{i_t} \mid t) \right],
    \label{eq:total_loss}
\end{equation}
can be regarded as an unbiased estimator, note the following:

\begin{equation}
    g(\theta, \epsilon, t) = \left( \epsilon_\theta(f_{i_t} \mid t) - \epsilon \right) \nabla_\theta \epsilon_\theta(f_{i_t} \mid t).
    \label{eq:uneq2}
\end{equation}

Then the gradient can be expressed as:
\begin{equation}
    \nabla_\theta \mathcal{L}_\theta(\mathbf{D}) = \frac{2}{n} \sum_{i=1}^n \mathbb{E}_{\epsilon, t}[g(\theta, \epsilon, t)].
    \label{eq:eq3}
\end{equation}
    
When approximating the expectation using Monte Carlo samples \(\{(\epsilon_j, t_j)\}_{j=1}^M\), the estimator becomes:
\begin{equation}
    \widehat{\nabla_\theta \mathcal{L}_\theta(\mathbf{D})} = \frac{2}{n} \sum_{i=1}^n \frac{1}{M}\sum_{j=1}^M g(\theta, \epsilon_j, t_j).
    \label{eq:eq4}
\end{equation}
    
Due to the \textbf{linearity of expectation}, we have:
\begin{equation}
    \mathbb{E}\left[\widehat{\nabla_\theta \mathcal{L}_\theta(\mathbf{D})}\right] = \frac{2}{n} \sum_{i=1}^n \mathbb{E}_{\epsilon, t}[g(\theta, \epsilon, t)] = \nabla_\theta \mathcal{L}_\theta(\mathbf{D}).
    \label{eq:eq5}
\end{equation}
    
Under appropriate regularity conditions (such as differentiability and integrability), the gradient operator can be interchanged with the expectation operator:
\begin{equation}
    \nabla_\theta \mathbb{E}_{\epsilon,t}[g(\theta, \epsilon, t)] = \mathbb{E}_{\epsilon,t}[\nabla_\theta g(\theta, \epsilon, t)].
    \label{eq:eq6}
\end{equation}

Therefore, the sample-based gradient computed in Eq.~\eqref{eq:total_loss} is an \emph{unbiased estimator} of the true gradient.


\end{document}